# Real-Time Drivers' Drowsiness Detection and Analysis through Deep Learning

1st ANK Zaman
*Member, IEEE*
https://orcid.org/0000-0001-7831-0955

2nd Prosenjit Chatterjee
*Dept of Computer Science & Cyber Security*
*Southern Utah University*
Ceder City, UT, USA
prosenjitchatterjee@suu.edu

3rd Rajat Sharma
*Dept of Physics & Computer Science*
*Wilfrid Laurier University*
Waterloo, ON, Canada
shar5100@mylaurier.ca

*Abstract*—A long road trip is fun for drivers. However, a long drive for days can be tedious for a driver to accommodate stringent deadlines to reach distant destinations. Such a scenario forces drivers to drive extra miles, utilizing extra hours daily without sufficient rest and breaks. Once a driver undergoes such a scenario, it occasionally triggers drowsiness during driving. Drowsiness in driving can be life-threatening to any individual and can affect other drivers' safety; therefore, a real-time detection system is needed. To identify fatigued facial characteristics in drivers and trigger the alarm immediately, this research develops a real-time driver drowsiness detection system utilizing deep convolutional neural networks (DCNNs) and OpenCV.Our proposed and implemented model takes real-time facial images of a driver using a live camera and utilizes a Python-based library named OpenCV to examine the facial images for facial landmarks like sufficient eye openings and yawn-like mouth movements. The DCNNs framework then gathers the data and utilizes a pre-trained model to detect the drowsiness of a driver using facial landmarks. If the driver is identified as drowsy, the system issues a continuous alert in real time, embedded in the Smart Car technology.By potentially saving innocent lives on the roadways, the proposed technique offers a non-invasive, inexpensive, and cost-effective way to identify drowsiness. Our proposed and implemented DCNNs embedded drowsiness detection model successfully react with NTHU-DDD dataset and Yawn-Eye-Dataset with drowsiness detection classification accuracy of 99.6% and 97% respectively.

## I. INTRODUCTION

According to research conducted by the AAA Foundation for Traffic Protection and Safety, an anticipated 328,000 drowsy driving accidents happen each year, which is over three times the number reported to police [1]. Among these incidents, approximately 109,000 resulted in injuries, while about 6,400 were fatal. The outcomes of this study reveal that the actual occurrence of fatalities due to drowsy driving may exceed reported figures by more than 350% – a recent report from the National Safety Council published in 2021. National Highway Traffic Safety Administration (NHTSA) [2] reported that "Fatigue-related crashes resulting in injury or death cost society $109 billion annually, not including property damage". Several studies [3], [4] investigated sleepiness detection systems that employ a variety of ways to address this issue. These techniques included EEG, ECG, and eye-tracking technologies. Unfortunately, these approaches are less appropriate for real-world applications because of their cost and inappropriateness [5].

In recent years, machine learning in computer vision [6] have shown promise in detecting driver fatigue [7]. Our experiment utilizes a Deep CNNs [8] algorithm and OpenCV [9] libraries to develop a real-time driver drowsiness detection system. The proposed model takes a set of facial images of the driver's faceial image setsusing a live sensor camera, utilizes OpenCV libraries to evaluate the driver's face images and finds facial landmarks like eye closings, mouth openings, and fatigued expressions for drowsiness identifiers. The applied algorithms in CNNs receives the data utilizing the pre-trained model for the purpose of detecting drowsiness through the deviation of facial features [10]

For a better road safety, the proposed and implemented system can play a pivotal role in detecting driver's drowsiness in real time.

Our experiments show promising results for two different datasets: i) a large dataset, 'NTHU-DDD [11],' with 99.6% classification accuracy and ii) a smaller dataset, 'Yawn-Eye-Dataset [12],' with 97% classification accuracy. This experiment provides a strategy to mitigate driver drowsiness and improve long distance driving safety.

Installation: Before running the code, download and install the required libraries, including OpenCV, TensorFlow, Keras, and Anaconda to run the Python notebook file in Jupyter Notebook.

Usage: The system utilizes a real-time live camera to capture facial images of the driver at regular intervals. It then processes these images using OpenCV libraries to detect facial traits. The CNNs model then analyze the facial features to determine if the driver is drowsy.

Functionality: The Driver Drowsiness Detection System analyzes the driver's facial features such as deviation in facial expressions, eye movements, and mouth openings to identify the signs of fatigue. When the drowsiness is detected, the system triggers a warning alert, helping to prevent accidents and improve road safety.

The rest of the paper is organized as follows: Section II discusses related work, Section III describes the datasets used in this research, Section IV details the technical aspects of the implemented models, Section V presents the experimental results, and Section VI provides the concluding remarks.

## II. RELATED WORK

Several studies have focused on developing driver drowsiness detection systems. Bio-signals, including electroencephalogram (EEG), electrocardiogram (ECG), etc were exploited to detect driver's physiological state [13], [14]. Whereas the driving patterns, eyes and facial movements [5] were commonly utilized where machine learning model were predominantly used [5], [15], [16], [17].

They analyzed facial features, embedded with attention mechanisms to focus on facial features such as eye movements and facial expressions [15]. They achieved a decent classification accuracy of 93.5% [15].

In another experiment Kumar, et. al. [16], proposed a driver drowsiness detection system, that includes face and eye features using machine learning algorithms. The algorithm used Support Vector Machines (SVM) as a classifier to detect the signs of intoxication [16].

Jebraeily et al. [17] propose a convolutional neural networks (CNNs) based technique for driver drowsiness detection using a genetic algorithm. In another experiment, the authors contributed a new driver drowsiness dataset based on the FER-2013 dataset [18], employing transfer learning to fine-tune the optimized CNNs for the drowsiness dataset and evaluating the proposed method against other CNNs architectures like ResNet, VGG, and GoogleNet. Their proposed method brings up 99.6% [17] classification accuracy on drivers drowsiness detection.

Bai et al. [19] represent a method for detecting driver drowsiness using spatial-temporal graph convolution networks (ST-GCNs). Their approach integrates spatial and temporal features from video sequences to enhance the accuracy of drowsiness detection [19]. They tested their method on two datasets with annotated drowsiness levels, demonstrating significant improvements over existing techniques [19]. The model's architecture utilizes two streams: one for spatial features extracted from individual frames and another for temporal features capturing motion patterns [19]. The dual-stream approach effectively captures the complex dynamics of drivers' behaviour, providing an adequate solution for real-time drowsiness monitoring in driving situations [19]. The proposed system achieved 97% [19] classification accuracy [19].

Convolutional Neural Networks (CNNs) was implemented in several deep learning-based experiments for the drivers' drowsiness detection [20], [21], utilizing facial expressions and eye movements, and achieved accuracy around 97% [20], [21] on an average.

## III. DATASET

In this research, we utilized two publicly available datasets named NTHU-DDD [11] from the Computer Vision Lab, National Tsing Hua University, Taiwan and Yawn-Eye-Dataset [12]. The NTHU-DDD [22] dataset consists of 66,521 images incorporating 36 subjects from various ethnic backgrounds. These subjects were recorded with and without glasses or sunglasses in a range of simulated driving scenarios. The scenarios include everyday driving, yawning, slow blinking, falling asleep, and bursting into laughter, under both day and night lighting conditions. The recordings were conducted while the subjects sat on a chair playing a basic driving simulation game using simulated steering wheels and pedals. During these sessions, they were guided to display a series of frontal facial expressions. The dataset has two different types of image sets, i) drowsy and ii) not drowsy categories containing 36,030 drowsy images and 30491 not drowsy images, respectively [22]. Yawn-Eye-Dataset [12] consists of 4 target classes, namely "Closed," "No Yawn," "Yawn," and "Open." Each category of the said classes was distributed among train and test sets. Each set has approximately 617 training images and approximately 109 test images. We used approximately 2400 training images and 400 test images to conduct our research. Figure 1 shows a driver in a normal state (open eye and conscious) and fatigued state (yawning and/or closed eye) taken from the dataset [12].

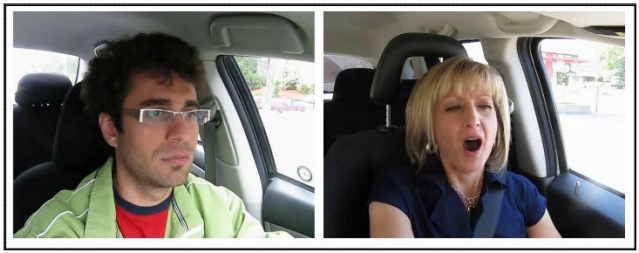

Fig. 1: A Normal and Drowsy Driver Scenario

## IV. MODEL DESCRIPTION

The convolutional neural networks (CNNs) model employed in this experiment comprises two convolutional layers followed by max-pool layers, a flattening layer, a dense (fully connected) layer, a dropout layer, and an output layer. There are 32 filters used by the first convolution layer with a 5x5 kernel size, that provides an output of a 94x94x32 feature map, followed by a 2x2 max-pool layer, reducing the dimensions to 47x47x32. The second convolutional layer applies 64 filters with a 3x3 kernel, resulting in a 45x45x64 features map, which is subsequently reduced to 22x22x64 by another 2x2 max-pool layer. The output is then flattened to a 1D tensor with 30,976 units, which feeds into a dense layer with 128 units. A dropout layer follows to reduce overfitting [23], [24], [25] with the final dense layer providing a 4-unit output for classification. It uses (Rectifier Linear Unit) ReLU activations in hidden layers Sigmoid (for 2-class classification) activation and SoftMax (for multi-class classification) activation in the output layer to predict the class probabilities for the 2-class (NTHU-DDD) and 4-class (Yawn-Eye-Dataset) classification tasks, respectively. Figure 2 visually represents our implemented DCNNs model. In the preprocessing steps, we resized the entire dataset images into 96x96 resolution for CNNs input. In our experiment, the implemented CNNs offer a practical, real-time approach for spotting drowsy drivers based on detecting the deviation of facial traits from normal to drowsy conditions. Our implemented system is available in [26].

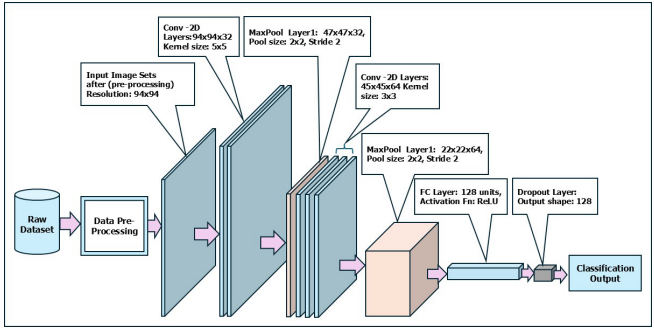

Fig. 2: DCNNs Framework Structure

TABLE I: DCNNs Model Description

| Sr. No. | Layer Name | Layer Structure Parameters Configuration | Significance |
|---|---|---|---|
| 1 | Input Image Processing | Resolution: 96x96 pixels | Prepares raw data for feature extraction |
| 2 | Convolutional Layer - Conv1 | Filters: 32, Kernel Size: 5x5 | Extracts basic features from input images |
| 3 | Max-Pooling Layer - MaxPool1 | Size: 2x2, Stride: 2x2 | Reduces spatial dimensions to decrease computation |
| 4 | Convolutional Layer - Conv2 | Filters: 64, Kernel Size: 3x3 | Extracts more complex features from previous layer outputs |
| 5 | Max-Pooling Layer - MaxPool2 | Size: 2x2, Stride: 2x2 | Further reduces spatial dimensions for computational efficiency |
| 6 | Fully Connected Layer - FC1 | Activation Units: 128 | Processes extracted features for high-level reasoning |
| 7 | Densc Layer | Activation Units: 128 | Continues high-level processing towards final output |
| 8 | Dropout Layer | Rate: 20% | Prevents overfitting by randomly dropping units during training |
| 9 | Output Classification Layer | Not Applicable | Produces final decision based on processed features |

## V. RESULT AND DISCUSSION

Our proposed and implemented DCNNs model features a lighter layer structure without compromising its robust classification capabilities. This streamlined, pretrained model enables quicker responses in real-time applications.

In our experiment, we train the DCNNs model with 30 iterations, with 78 images per batch cycle in each epoch. Our finding shows the robustness of our proposed and implemented system that works accurately to identify drivers' drowsiness depending on drivers' facial traits. Figures 6, and 9 show the evaluation matrices of DCNNs, on 'NTHU-DDD' dataset and 'Yawn-Eye-Dataset' respectively. We achieved a promising 99.6% test accuracy on 'NTHU-DDD' dataset and 97% test accuracy on 'Yawn-Eye-Dataset'. Figure 3 shows the real-time drowsiness detection through the DCNNs model using real-time feed from a webcam to OpenCV. In figure 3a, 3b, and 3c our system triggers warning signal to the driver when the subject-driver exhibits either or all of the facial expressions like eyes-closed and yawning. Figure 3d will not trigger any alert as subject driver looks alert.

### A. Evaluation Metrics

Precision is the ratio of true positives (TP) to the sum of true positives and false positives (FP). Recall is the ratio of true positives (TP) to the sum of true positives and false negatives (FN). f1-Score is the harmonic mean of precision and recall and is used as a single metric that balances the trade-off between precision and recall. The F1score ranges

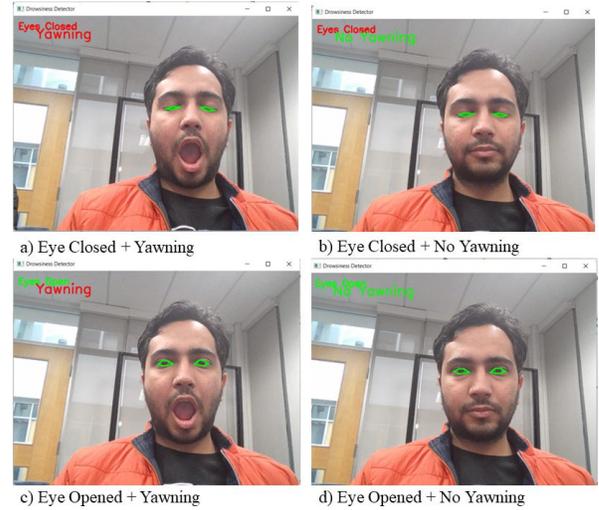

a) Eye Closed + Yawning   b) Eye Closed + No Yawning
c) Eye Opened + Yawning   d) Eye Opened + No Yawning

Fig. 3: Real-Time Drowsiness Detection System Outputs

from 0 to 1, with 1 being the best possible score. The formal definitions of the mentioned terms are given below:

$Accuracy = \frac{TP+TN}{TP+TN+FP+FN}$

Here in this case, Precision is the ratio of true positives (TP) to the sum of true positives and false positives (FP).

$Precision = \frac{TP}{TP+FP}$

Recall: This is the ratio of true positives (TP) to the sum of true positives and false negatives (FN).

$Recall = \frac{TP}{TP+FN}$

f1 score is the harmonic mean of precision and recall and is used as a single metric that balances the trade-off between precision and recall. The f1 score ranges from 0 to 1, with 1 being the best possible score.

$f1 = \frac{2*Precision*Recall}{Precision+Recall} \quad \frac{2*TP}{2*TP+FP+FN}$

### B. Results on NTHU-DDD

Figure 4, 5, and 6 show accuracy vs. loss graph for 30 epochs, confusion matrix, and the significance of the DCNNs implementation as we obtained 99.6% test accuracy on NTHU-DDD dataset. We observed a steady improvement in training accuracy over the epochs, as shown in Figure 4. The confusion matrix in Figure 5 indicates a minimal number of true negatives (27) and false positives (21). Figure 6 highlights the model's strong performance, demonstrating promising precision, recall, and F1-score. Additionally, under the evaluation metrics shown in Figure 6, the support indicates no data loss, confirming the DCNNs model's robust performance on the NTHU-DDD dataset.

### C. Results on Yawn-Eye-Dataset

Figure 7, 8, and 9 show accuracy vs. loss graph for 30 epochs, confusion matrix, and the significance of the DCNNs implementation as we obtained 97% test accuracy on Yawn-Eye-Dataset. As illustrated in Figure 7, the model's validation accuracy on the Yawn Eye dataset exceeds 90% after approximately 15 epochs, while the training accuracy steadily improves, approaching nearly 100% after the same period. The confusion matrix in Figure 8 highlights the model's strong

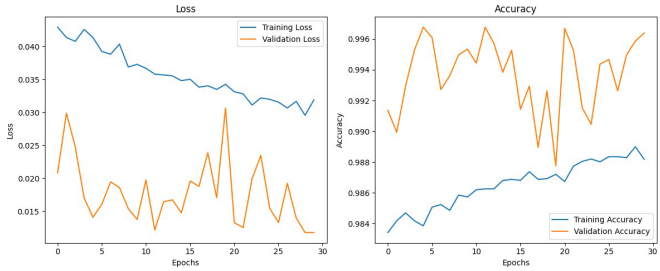

Fig. 4: Accuracy vs. Loss Graph for NTHU-DDD Dataset

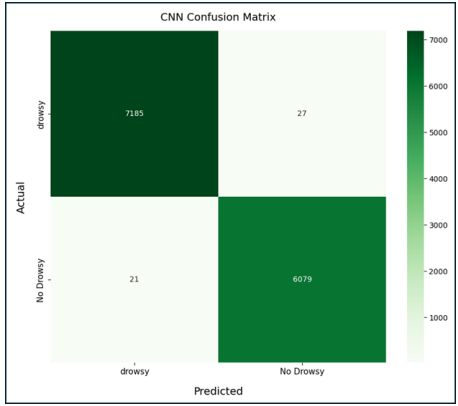

Fig. 5: Confusion Matrix for the NTHU-DDD Dataset

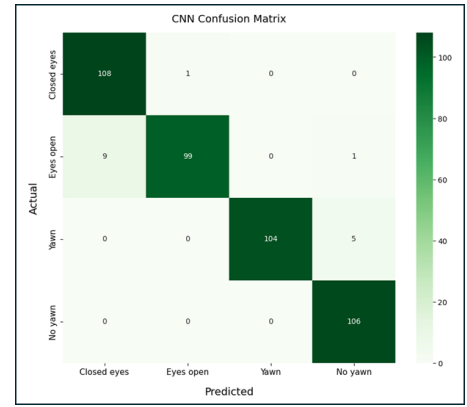

Fig. 8: Confusion Matrix for the Yawn-Eye-Dataset

|  | precision | recall | f1-score | support |
|---|---|---|---|---|
| closed eyes | 0.92 | 0.99 | 0.96 | 109 |
| eyes open | 0.99 | 0.91 | 0.95 | 109 |
| yawn | 1.00 | 0.95 | 0.98 | 109 |
| no yawn | 0.95 | 1.00 | 0.97 | 106 |
| accuracy |  |  | 0.96 | 433 |
| macro avg | 0.96 | 0.96 | 0.96 | 433 |
| weighted avg | 0.97 | 0.96 | 0.96 | 433 |

Fig. 9: Precision, recall, f1-score and support on Test Data for Yawn-Eye-Dataset

performance in a multiclass environment. In the evaluation matrices shown in Figure 9, the support confirms no data loss in the Yawn-Eye-Dataset while implementing the same DCNNs model. The finding demonstrated a reliable real-time drivers drowsiness detection system with a low rate of false alarm. It is evident that the dataset samples were collected using inside car camera utilizing various real time lighting condition and drivers' position to help developing a robust drowsiness detection system.

Table II demonstrates the comparison study of classification accuracies of the implemented model with the recently published works in this field of research. The research findings show that the proposed Driver Drowsiness Detection System successfully identifies driver sleepiness based on facial features. The framework can increase traffic safety through a warning sound to make the drivers awake while driving.

|  | precision | recall | f1-score | support |
|---|---|---|---|---|
| drowsy | 1.00 | 1.00 | 1.00 | 7212 |
| notdrowsy | 1.00 | 1.00 | 1.00 | 6100 |
| accuracy |  |  | 1.00 | 13312 |
| macro avg | 1.00 | 1.00 | 1.00 | 13312 |
| weighted avg | 1.00 | 1.00 | 1.00 | 13312 |

Fig. 6: Precision, recall, f1-score and support on Test Data for NTHU-DDD Dataset

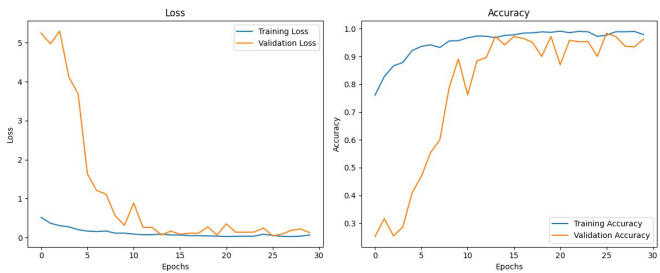

Fig. 7: Accuracy vs. Loss Graph for Yawn-Eye-Dataset

## VI. CONCLUSION

Our proposed and implemented Driver Drowsiness Detection System uses DCNNs and OpenCV frameworks, and utilizes facial features like eye opening or closing, and facial movements like yawning or no-yawn, to identify driver drowsiness for a low-cost effective solution. We achieved a promising 99.6% test accuracy and 97% test accuracy for 'NTHU-DDD' and 'Yawn-Eye-Dataset', respectively, which is significantly higher compared to the recently implemented DCNNs performance under the Computer vision domain mentioned in Table II. The frameworks detects drivers drowsiness in real-time with promising accuracy, precision, and very few false alarms. Our proposed and implemented frameworks can be embedded either as a standalone or integrated into the current car system. The device can increase traffic safety by sending warning sounds to drowsy drivers and waking them while driving, or they can take precautionary measures thereafter. The system depends on a live sensor camera to take facial images in real-time framework implementation. It might not function well in very low luminance or if the facial sensor capture device is restricted. To obtain better precision and robustness of the system, further research can be conducted incorporating infrared camera sensors or the inclusion of other data sources for the multimodal performance comparison and fusions.

| Reference | Model | Dataset | Accuracy |
|---|---|---|---|
| Bai et. Al. [19] (2022) | ST-CNNs | NTHU-DDD | 92.7% |
| Ed-Doughmi et al. [27] (2020) | RNN | NTHU-DDD | 92.00% |
| Dua et al. [28] (2020/2021) | Deep CNN | NTHU-DDD | 85.00% |
| Saif and Mahayuddin [29] (2020) | DCNN | iBUG 300 W | 98.97% |
| Quddus et al. [30] (2021) | LSTM, CNN | 38 drivers simulation | 95% - 97% |
| Wijnands et al. [31] (2019) | 3D neural networks | DDD | 80.8% |
| Vijaypriya & Mohan [32] (2023) | Res-Net | NTHU-DDD | 96.14% |
| **Proposed Model on NTHU-DDD** | **DCNNs** | **NTHU-DDD** | **99.6%** |
| **Proposed Model on Yawn-Eye-Dataset** | **DCNNs** | **Yawn-Eye-Dataset** | **97%** |

TABLE II: Summary of different studies and their accuracies with the proposed model

To increase the precision and dependability of the system, more investigation can be done, such as incorporating infrared camera sensors or the inclusion of other data sources for the multimodal performance comparison and fusions.